\documentclass[pmlr]{jmlr}

\usepackage{bbm}
\usepackage{bigstrut}
\usepackage{booktabs}
\usepackage[font=small,labelfont=bf,format=hang]{caption}
\usepackage{color}
\usepackage{float}
\usepackage{longtable}
\usepackage{mathrsfs}
\usepackage{multirow}
\usepackage{nccmath}
\usepackage{placeins}
\usepackage{setspace}
\usepackage{subcaption}
\usepackage[load-configurations=version-1]{siunitx} 
\usepackage{verbatim}

\firstpageno{1}

\jmlrvolume{85}
\jmlryear{2018}
\jmlrworkshop{Machine Learning for Healthcare}

\title[Bidirectional Language Modeling for Transfer Learning in Biomedical NER]{Effective Use of Bidirectional Language Modeling for Transfer Learning in Biomedical Named Entity Recognition}

\author{\Name{Devendra Singh Sachan}$^{\spadesuit}$ \Email{devendra.singh@petuum.com}\\
\Name{Pengtao Xie}$^{\spadesuit}$ \Email{pengtao.xie@petuum.com}\\ 
\Name{Mrinmaya Sachan}$^{\triangle}$ \Email{mrinmays@cs.cmu.com}\\
\Name{Eric P. Xing}$^{\spadesuit}$ \Email{eric.xing@petuum.com}\\
\addr $^{\spadesuit}$Petuum Inc, Pittsburgh, PA, USA\\
$^{\triangle}$Machine Learning Department, CMU, Pittsburgh, PA, USA}

\editor{...}

\begin{document}
\maketitle

\begin{abstract}%
Biomedical named entity recognition (NER) is a fundamental task in text mining of medical documents and has many applications. Deep learning based approaches to this task have been gaining increasing attention in recent years as their parameters can be learned end-to-end without the need for hand-engineered features. However, these approaches rely on high-quality labeled data, which is expensive to obtain. To address this issue, we investigate how to use unlabeled text data to improve the performance of NER models. Specifically, we train a bidirectional language model (BiLM) on unlabeled data and transfer its weights to ``\textit{pretrain}" an NER model with the same architecture as the BiLM, which results in a better parameter initialization of the NER model. We evaluate our approach on four benchmark datasets for biomedical NER and show that it leads to a substantial improvement in the F1 scores compared with the state-of-the-art approaches. We also show that BiLM weight transfer leads to a faster model training and the pretrained model requires fewer training examples to achieve a particular F1 score.
\end{abstract}


\flushbottom

\section{Introduction}
\begin{figure}[t]
\centering
\includegraphics[width=0.7\linewidth]{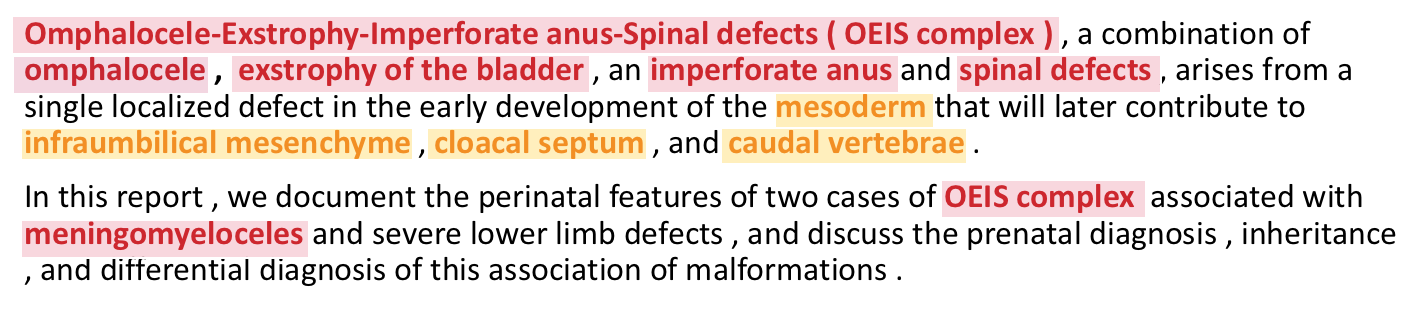}
\caption{Example of disease and anatomical entities in medical text. Disease entities are highlighted in red and anatomical entities are highlighted in yellow.}
\label{fig:ner_example}
\end{figure}

The field of biomedical text mining has received increased attention in recent years due to the rapid increase in the number of publications, scientific articles, reports, medical records, etc.\ that are available and readily accessible in electronic format. These biomedical data contains many mentions of biological and medical entities such as chemical ingredients, genes, proteins, medications, diseases, symptoms, etc. Figure~\ref{fig:ner_example} shows a medical text that contains seven disease entities (highlighted in red) and four anatomical entities (highlighted in yellow). The accurate identification of such entities in text collections is a very important subtask for information extraction systems in the field of biomedical text mining as it helps in transforming the unstructured information in texts into structured data. Search engines can index, organize, and link medical documents using such identified entities and this can improve medical information access as the users will be able to gather information from many pieces of text. The identification of entities can also be used to mine relations and extract associations from the medical research literature, which can be used in the construction of medical knowledge graphs~\citep{doi:10.1038/s41598-017-05778-z}. We refer to this task of identification and tagging of entities in a text as members of predefined categories such as diseases, chemicals, genes, etc.\ as named entity recognition (NER).

NER has been a widely studied task in the area of natural language processing (NLP) and a number of works have applied machine learning approaches to NER in the medical domain. Building NER systems with high precision and high recall for the medical domain is quite a challenging task due to high linguistic variation in data. First, a dictionary-based approach doing pattern matching will fail to correctly tag ambiguous abbreviations that can belong to different entity types. For example, the term CAT can refer to several phrases\textemdash``\emph{chloramphenicol acetyl transferase}," ``\emph{computer-automated tomography}," ``\emph{choline acetyltransferase}," or ``\emph{computed axial tomography}"~\citep{stevenson2010disambiguation}. Second, as the vocabulary of biomedical entities such as proteins is quite vast and is rapidly evolving, it makes the task of entity identification even more challenging and error-prone as it is difficult to create labeled training examples having a wide coverage. Also, in contrast to natural text, entities in the medical domain can have very long names as shown in Figure~\ref{fig:ner_example} that can lead an NER tagger to incorrectly predict the tags. Lastly, state-of-the-art machine learning approaches for NER task rely on high-quality labeled data, which is expensive to procure and is therefore available only in limited quantity. Therefore, there is a need for approaches that can use unlabeled data to improve the performance of NER systems.

NER can be devised as a supervised machine learning task in which the training data consists of labels or tags for each token in the text. A typical approach for NER task is to extract word-level features followed by training a linear model for tag classification. To extract features, our NER model makes use of pretrained word embeddings, learned character features, and word-level bidirectional long short-term memory (BiLSTM). The word embeddings are learned from a large collection of PubMed abstracts and it improves the F1 score on NER datasets compared with randomly initialized word vectors. The BiLSTM effectively models the left and right context information around the center word for every time step and this context based representation of a word can help in the disambiguation of abbreviations. The BiLSTM, when applied in combination with character features also maps similar terms like ``\emph{lymphoblastic leukemia}," ``\emph{null-cell leukemia}," and its varied forms in a latent space that captures the semantic meaning in the phrases. This powerful representation of terms in a latent semantic space can also help in the correct classification of unseen entities as entities with similar contexts are mapped closer together. 

In this paper, we propose a transfer learning approach that makes use of unlabeled data to pretrain the weights of NER model using an auxiliary task. Specifically, we do language modeling in both the forward and backward directions to pretrain the weights of NER model that is later fine-tuned using the supervised task-specific training data. We believe that such generative model pretraining can prevent overfitting, improve model training, and its convergence speed. We show that such pretraining of weights helps to substantially increase the F1 score on four benchmark datasets for biomedical NER compared with the state-of-the-art approaches. We also observe that BiLM weight transfer leads to faster convergence during the NER model fine-tuning step. As an unsupervised method, our transfer learning approach requires only unlabeled data and thus is generally applicable to different NER datasets compared with the supervised transfer learning approaches that rely on task-specific labeled data to pretrain the model parameters~\citep{lee18transfer}.

Following this Introduction, the remainder of this paper is organized as follows. Section~\ref{sec:methods} explains the NER model, its training methodology, and bidirectional language modeling. Section~\ref{sec:exp_setup} describes the experimental setup such as datasets, model architecture, and the training process. Section~\ref{sec:results} reports the results on these datasets and analyzes the performance of the pretrained NER model in detail. Section~\ref{sec:related_word} reviews the related work for biomedical NER. The conclusion, in section~\ref{sec:conclusion}, summarizes our methods, results, and discusses the future work.

\section{Methods} \label{sec:methods}
\begin{figure}[t]
\centering
\begin{minipage}{.5\textwidth}
  \centering
  \includegraphics[scale=0.45]{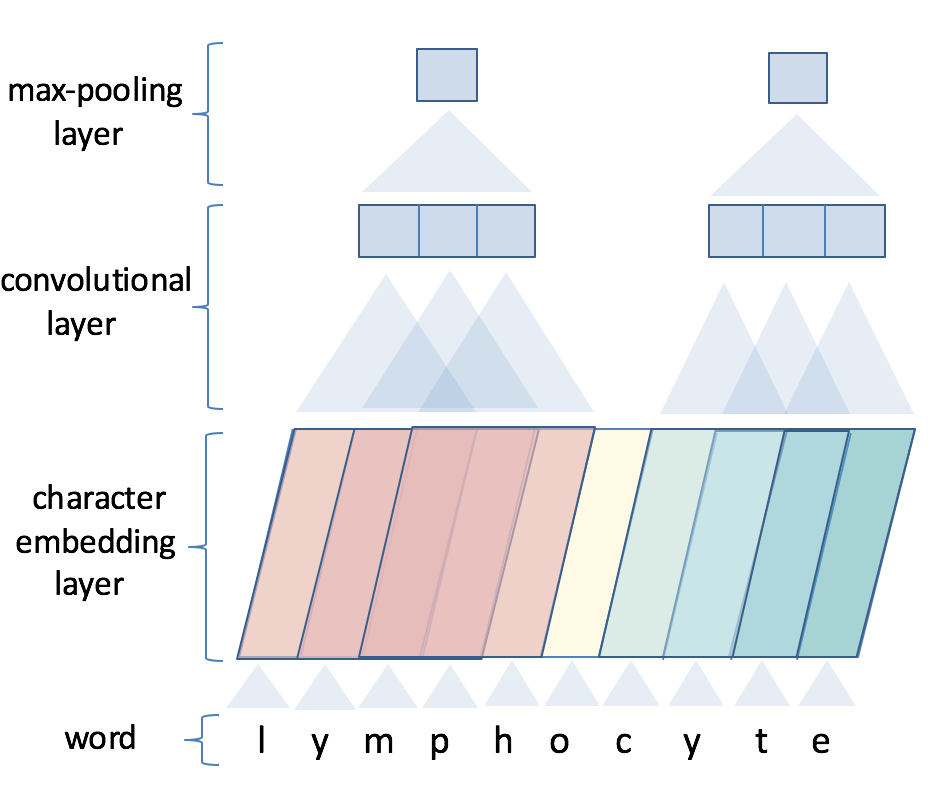}
  \caption{Character CNN block diagram}
  \label{fig:cnn}
\end{minipage}%
\begin{minipage}{.5\textwidth}
  \centering
  \includegraphics[scale=0.60]{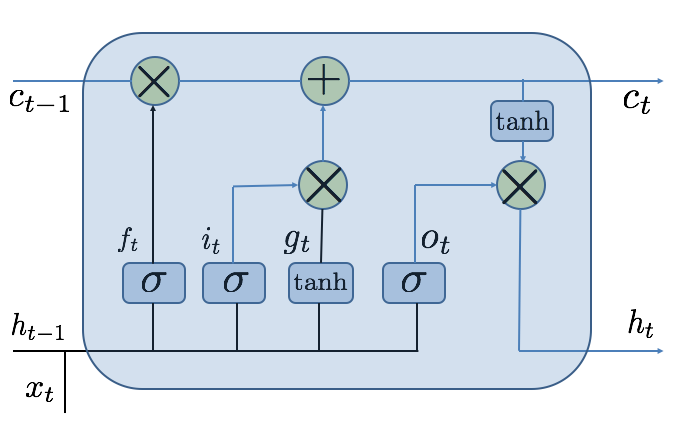}
  \caption{LSTM block diagram}
  \label{fig:lstm}
\end{minipage}
\end{figure}
The main building blocks of our neural network based NER model are: character-level convolutional neural network (CNN) layer, word embedding layer, word-level BiLSTM layer, decoder layer, and sentence-level label prediction layer (see Figure \ref{fig:ner_model}). During model training, all the layer are jointly trained. Before training, we also pretrain the parameters of the character-CNN, word embedding, and BiLSTM layers in the NER model using the learned parameters from a language model that has the same architecture. Specifically, we perform bidirectional language modeling (BiLM) to pretrain the weights of both the forward and backward LSTMs in the NER model. Next, we will describe these layers in detail.

\subsection{Character-Level CNN} CNNs~\citep{lecun1990handwritten} are widely used in computer vision tasks for visual feature extraction~\citep{krizhevsky2012imagenet}. In NLP, where the data is mostly sequential, successful applications of CNNs include tasks such as text classification~\citep{kim2014convolutional} and sequence labeling~\citep{collobert2011natural}. In this paper, we use CNNs to extract features from characters~\citep{kim2016character} as they can encode morphological and lexical patterns observed in languages.

Similar to the concept of word embedding, each character is represented by an embedding vector. These character embeddings are stored in a lookup table $W_c \in \mathbb{R}^{V_c \times D_c}$, where $V_c$ is the character vocabulary, $D_c$ is the dimensionality of character embeddings. To compute character-level features, we perform 1D convolution along the temporal dimension.\footnote{To have a uniform length, each word is right-padded with a special padding token so that the length of every word is the same as that of the longest word in every mini-batch. The embedding of the padding character is always a zero vector.} Mathematically, this can be written as:
\begin{align*}
z_{k}[i] = f(W_{k} * X[:,\;i + s - 1] + b_k),
\end{align*}
where $*$ is the dot product operator, $b_k$ is the bias, $X \in \mathbb{R}^{D_c \times w_{\ell}}$ is the character-based embedding representation of a word, $w_{\ell}$ is the length of a word, $W_k$ are filter weights, $s$ is the convolution stride, $f$ can be any nonlinear function such as $\tanh$ or rectified linear units ($f(x) = \max(0,\:x)$). To capture the important features of a word, multiple filters of different strides are used. Finally, the maximum value is computed over the time dimension also called \emph{max-pooling} to get a single feature for every filter weight. All the features are concatenated to obtain character-based word representation $v_{\textit{char}}^{\textit{w}}$. A block diagram of character-level CNN is shown in Figure~\ref{fig:cnn}.

\subsection{Word-Level Bidirectional LSTM}
Recurrent neural network~\citep{werbos1988generalization} such as LSTM~\citep{Hochreiter:1997:LSM:1246443.1246450} is widely used in NLP because it can model the long-range dependencies in language structure with their memory cells and explicit gating mechanism. The dynamics of an LSTM cell is controlled by an input vector ($x_t$), a forget gate ($f_t$), an input gate ($i_t$), an output gate ($o_t$), a cell state ($c_t$), and a hidden state ($h_t$), which are computed as:
\begin{align*}
i_t &= \sigma(W_i * [h_{t-1}, x_t] + b_i) \\
f_t &= \sigma(W_f * [h_{t-1}, x_t] + b_f) \\
o_t &= \sigma(W_o * [h_{t-1}, x_t] + b_o) \\
g_t &= \tanh(W_g * [h_{t-1}, x_t] + b_g) \\
c_t &= f_t \odot c_{t-1} + i_t \odot g_t \\
h_t &= o_t \odot \tanh(c_t),
\end{align*}
where $c_{t-1}$ and $h_{t-1}$ are the cell state and hidden state respectively from previous time step, $\sigma$ is the sigmoid function ($\frac{1}{1+e^{-x}}$), $\tanh$ is the hyperbolic tangent function ($\frac{e^{x}-e^{-x}}{e^{x}+e^{-x}}$), $\odot$ denotes element-wise multiplication. Figure~\ref{fig:lstm} shows the block diagram of an LSTM cell. The parameters of the LSTM are shared for all the time steps. 

In our NER model, the word embeddings ($v^{w}_{\textit{emb}}$) and the character CNN features of a word ($v^{w}_{\textit{char}}$) are concatenated and is given as input to the sequence encoder ($x_t = [v^{w}_{\textit{emb}}, v^{w}_{\textit{char}}])$. The sequence encoder consists of a forward LSTM and a backward LSTM, which is also known as bidirectional LSTM (BiLSTM)~\citep{schuster1997bidirectional}. The input to the backward LSTM cell is the reversed order of words in the sequence. 

\subsection{Word-Level Likelihood}
\begin{figure}[t]
\centering
\begin{minipage}{.5\linewidth}
  \centering
  \includegraphics[scale=1.0]{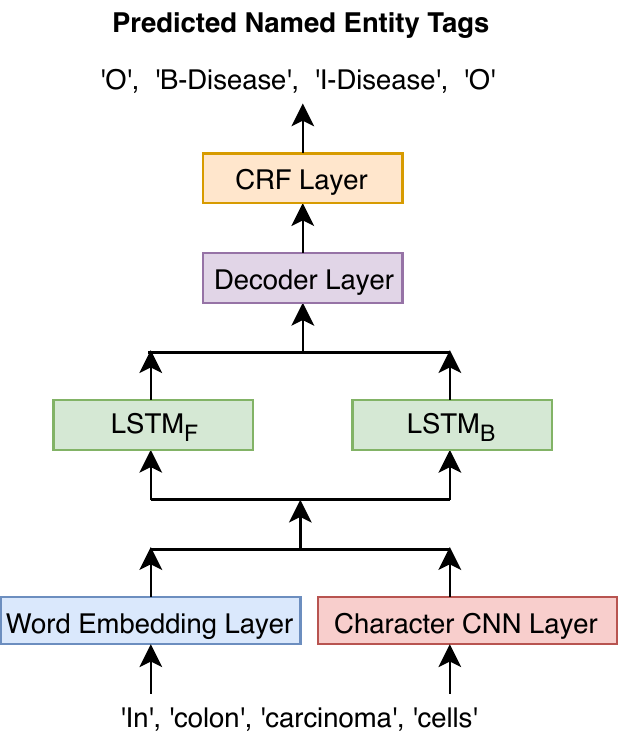}
  \caption{NER model architecture.}
  \label{fig:ner_model}
\end{minipage}%
\begin{minipage}{.5\linewidth}
  \centering
  \includegraphics[scale=1.0]{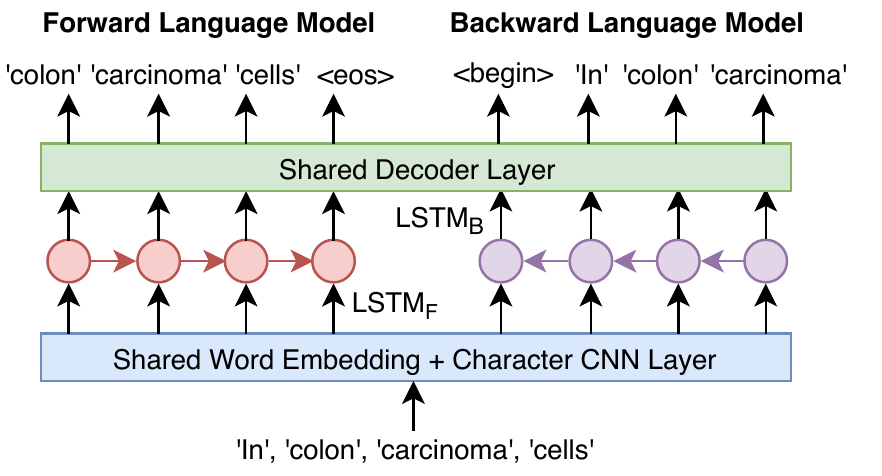}
  \caption{BiLM architecture.}
  \label{fig:lm_model}
\end{minipage}
\end{figure}

For every word, hidden state representations from BiLSTM are concatenated ($h_t = [\overrightarrow{h_t}, \overleftarrow{h_t}]$) and are fed to the decoder layer. The decoder layer computes an affine transformation of the hidden states
\begin{align*}
d_t = W_dh_t + b,
\end{align*}
where $H$ is the dimensionality of the BiLSTM hidden states, $T$ is the total number of tags, $W_d \in \mathbb{R}^{T\times H}$ and $b$ are learnable parameters. Decoder outputs are referred to as \textit{logits} in the subsequent discussions. To compute the probability of a tag ($\hat{y_t}$) for a word, softmax function is used
\begin{align*}
p(\hat{y}_t=j \mid w_t) = \mathrm{softmax}(d_t).
\end{align*}
Let $\mathbf{y} = \lbrace y_1, y_2,\ldots, y_N \rbrace$ denote the sequence of tags in the training corpus, then the cross-entropy loss is calculated as:
\begin{align*}
\mathrm{CE}_{\textit{ner}}(\mathbf{y}, \mathbf{\hat{y}}) = -\sum_{t=1}^{N} \sum_{j=1}^{T}\mathbbm{1}(y_t=\hat{y}_{t,j})\log{\hat{y}_{t,j}}
\end{align*}
Figure~\ref{fig:ner_model} shows the block diagram of our NER model. To learn the model parameters, average cross-entropy loss is minimized by backpropagation through time (BPTT)~\citep{werbos1990backpropagation}. When the word-level likelihood is minimized to train the NER model, it is denoted as CNN-BiLSTM.

\subsection{Sentence-Level Likelihood} A drawback of optimizing word-level likelihood is that it ignores the dependencies between other neighboring tags in the sentence. A better strategy is to model the entire sentence structure using a conditional random field (CRF). A CRF is a log-linear graphical model~\citep{Lafferty:2001:CRF:645530.655813} that additionally considers the transition score from one tag to the next tag. This encourages valid transition paths among the tags based on the learned transition parameters ($W^{\textit{crf}}\in\mathbb{R}^{T \times T})$. During training, we maximize the log-likelihood for the entire sentence.\footnote{This can be done in polynomial time using the \textit{forward-backward} algorithm~\citep[see][]{collins2013forward}.} Mathematically, this can be written as:\footnote{For a detailed derivation of CRF, see~\cite{collins2013crf}.}
\begin{align*}
\log p(\mathbf{y} \mid \mathbf{d}) = s(\mathbf{d}, \mathbf{y}) - \log\sum_{\mathbf{y'}\in S^m}e^{s(\mathbf{d}, \mathbf{y'})},
\end{align*}
where $S^m$ is the set containing all possible tag combinations for a sentence, $s(\mathbf{d}, \mathbf{y})$ is a scoring function defined as:
\begin{align*}
s(\mathbf{d}, \mathbf{y}) = \sum_{t=1}^{N-1} W_{y_t,y_{t+1}}^{\textit{crf}} + \sum_{t=1}^N d_{t,y_t}.
\end{align*}
In this paper, when the logits $d_{t,y_t}$ are fed to the CRF layer to optimize sentence likelihood, we call the NER model as CNN-BiLSTM-CRF. During inference, we use the Viterbi algorithm~\citep{forney1973viterbi} to find the best tag sequence that maximizes the sentence likelihood.

\subsection{Language Modeling}
Here, we provide a short description of language modeling, as its parameters are used to initialize the NER model.  In language modeling, the task is to train a model that maximizes the likelihood of a given sequence of words. At every step, a language model computes the probability of the next word in the sequence given all the previous words. If the sequence of words is $w_1, w_2,\ldots, w_n$, its likelihood is given as
\begin{align*}
p_f(w_1, w_2,\ldots, w_n) = \prod\limits_{i=2}^{n+1} p(w_i \mid w_1,\ldots,w_{i-1}),
\end{align*}
where $w_{n+1}$ is a special symbol for the end of a sequence. LSTM can be used to predict the probability of the next word given the current word and the previous sequence of words~\citep{graves2013generating}. This is done by applying an affine transformation to the hidden states of LSTM at every time step to obtain the logits for all the words in the vocabulary. We refer to this approach as the \emph{forward language model} ($LM_f$).

We can also model the reversed sequence of words in a similar manner. In this, we compute the probability of the reversed sequence as:
\begin{align*}
p_b(w_n, w_{n-1},\ldots, w_1) = \prod\limits_{i=n-1}^{i=0} p(w_i \mid w_{i+1},\ldots,w_n),
\end{align*}
where $w_0$ is a special symbol for the start of the sequence. We refer to this approach as the \emph{backward language model} ($LM_b$). The network architecture of both $LM_f$ and $LM_b$ is similar to the NER model (see Figure~\ref{fig:lm_model}). While training, both $LM_f$ and $LM_b$ share the parameters of the word embedding layer, character embedding layer, character CNN filters, and the decoder layer. We refer to this as the \emph{bidirectional language model} (BiLM). To learn the parameters of the BiLM, we perform joint training by minimizing the average cross-entropy losses of both the forward and backward language models
\begin{align*}
\mathrm{CE}_{\ell\textit{m}} = -\lambda_{\ell\textit{m}}(\log p_f(\mathbf{w}_{1:n}) + \log p_b(\mathbf{w}_{n:1})).
\end{align*}

\section{Experimental Setup}\label{sec:exp_setup}
In this section, we will first describe the datasets, their preprocessing, and performance evaluation criteria. Next, we will discuss the architecture of NER model and language model followed by their training details. 

\renewcommand{\thefootnote}{\Roman{footnote}}
\subsection{Dataset Preparation and Evaluation} 
We evaluate our proposed approach on four datasets: NCBI-disease~\citep{dougan2012improved}, BioCreative V Chemical Disease Relation Extraction (BC5CDR) task~\citep{li2016biocreative}, BC2GM~\citep{Ando2007BioCreativeIG}, and JNLPBA~\citep{kim2004introduction}. For each dataset, we use the training, development, and test set splits according to~\citet{crichton2017neural}.\footnote{For our experiments, we use the datasets publicly available at~\url{https://github.com/cambridgeltl/MTL-Bioinformatics-2016}.} An overall summary of these datasets such as the number of sentences, words, and entities is presented in Table~\ref{table:dataset_statistics}. For each dataset, we use its training and development splits as unlabeled data for language modeling task.

\begin{table}[t]
\centering
\begin{tabular}{l | r r r r}
\toprule
\textbf{Properties} & \textbf{NCBI-disease} & \textbf{BC5CDR} & \textbf{BC2GM} & \textbf{JNLPBA} \\
\midrule
Entity type & Disease & Disease, Chemical & Gene/Protein & 5 NEs \\
\# Entity mentions & 6,892 & 5,818 & 24,583  & 51,301 \\
\# Sentences & 7,295 & 13,907 & 20,000 & 24,806 \\
\# Words & 184,552 & 360,315 & 1,139,824 & 568,786 \\
\# Train documents & 593 & 500 & - & 1,800 \\
\# Dev documents & 100 & 500 & - & 200 \\
\# Test documents  & 100 & 500 & - & 404 \\
\bottomrule
\end{tabular}
\captionof{table}{General statistics of the datasets used in this work. `\#' symbol stands for the term `number of'. Entity types in JNLPBA dataset consists of protein, DNA, RNA, cell-type, and cell-line.}
\label{table:dataset_statistics}
\end{table}


We use a special token for the numbers and preserve case information. In all our experiments, we use IOBES tagging format~\citep{Borthwick:1999:MEA:930095} for the output tags. For evaluation, we report the precision, recall, and F1 scores for all the entities in the test set. We do exact matching of entity chunks to compute these metrics. For each dataset, we tune the hyperparameters of our model on the development set. Final training is done on both the training and development sets. We use PyTorch framework~\citep{paszke2017automatic} for all our experiments.

\subsection{Model Architecture Details} As we use language model weights to initialize the parameters of the NER model, both the models have identical configurations except for the top decoder layer. Dimensions of character embeddings and word embeddings are set to 50 and 300 respectively. CNN filters have widths ($w$) in the range from 1 to 7. The number of filters are computed as a function of filter width as $\mathrm{min}(200,\;50*w)$. The hidden state of LSTM has 256 dimensions. As the decoder layer is not shared between NER model and language model, the dimensions of the decoder layer are different for each of them. For NER model, as it concatenates the hidden states of forward and backward LSTM to give input to the decoder layer, the dimensions of the decoder matrix are $W_d^{\textit{ner}} \in \mathbb{R}^{512 \times T}$. For language model, dimensions of the decoder matrix are $W_d^{\ell\textit{m}} \in \mathbb{R}^{256 \times V}$ where $V$ is the vocabulary size.

\subsection{Language Model Training} We initialize the weights of word embedding layer for the BiLM task using pretrained word vectors. These vectors were learned using skip-gram~\citep{mikolov2013distributed} method\footnote{We learn word embeddings using the \emph{word2vec} toolkit: \url{https://code.google.com/p/word2vec/}} applied to a large collection of PubMed abstracts.\footnote{These PubMed abstracts are available from the BioASQ Task 4a challenge~\citep{283}.} The embeddings of out-of-vocabulary words are uniformly initialized. LSTM parameters are also uniformly initialized in the range $(-0.005, 0.005)$. For all the other model parameters, we use Xavier initialization~\citep{pmlr-v9-glorot10a}. 

For model training, we use mini-batch SGD with a dynamic batch size of 500 words. At the start of every mini-batch step, the LSTM starts from zero initial states. We do sentence-level language modeling and the network is trained using BPTT. We use Adam optimizer~\citep{kingma2014adam} with default parameters settings and decay the learning rate by 0.5 when the model's performance plateaus. We train the model for 20 epochs and do early stopping if the perplexity doesn't improve for 3 consecutive epochs. To regularize the model, we apply dropout~\citep{srivastava2014dropout} with probability $0.5$ on the word embeddings and LSTM's hidden states. To prevent the gradient explosion problem, we do gradient clipping by constraining its $L_2$ norm to be less than $1.0$~\citep{pascanu2013difficulty}.

\subsection{NER Model Training} To pretrain the NER model, we remove the top decoder layer of the BiLM and transfer the remaining weights to the NER model with the same architecture. Next, we fine-tune the pretrained NER model using Adam optimizer~\citep{kingma2014adam}. In contrast to random initialization, during fine-tuning, the pretrained weights act as the starting point for the optimizer. We use mini-batch SGD with a dynamic batch size of $1,000$ words and train the model for 50 epochs. Other settings are similar to the language model training procedure as described above.

\section{Results}\label{sec:results}
In this section, we first evaluate the BiLM pretrained NER model on four biomedical datasets and compare the results with the state-of-the-art models. Next, we analyze different variations of NER model pretraining and also do three experiments to study the properties of pretrained NER model. Finally, in a case study on NCBI-disease dataset, we analyze the model's predictions on disease entities. We use the CNN-BiLSTM-CRF architecture for NER model in all our the experiments unless specified otherwise.

\begin{table}[t]
\small
\centering
\begin{tabular}{l l r r r r r}
\toprule
\textbf{Dataset} & \textbf{Metric} & \textbf{Benchmark} & 
\textbf{FFN}
& 
\textbf{BiLSTM}
& \textbf{MTM-CW} & \textbf{BiLM-NER} \\
\midrule
\multirow{3}{*}{NCBI-disease} & Precision & 85.10 & - & 86.11 & 85.86 & \textbf{86.41} \\
& Recall & 80.80 & - & 85.49 & 86.42 & \textbf{88.31} \\
& F1 & 82.90 & 80.46 & 85.80 & 86.14 & \textbf{87.34} \\
\midrule
\multirow{3}{*}{BC5CDR} & Precision & 89.21 & - & 87.60 & \textbf{89.10} & 88.10 \\
& Recall & 84.45 & - & 86.25 & 88.47 & \textbf{90.49} \\
& F1 & 86.76 & 83.90 & 86.92 & 88.78 & \textbf{89.28} \\
\midrule
\multirow{3}{*}{BC2GM} & Precision & - & - & 81.57 & \textbf{82.10} & 81.81 \\
& Recall & - & - & 79.48 & 79.42 & \textbf{81.57} \\
& F1 & - & 73.17 & 80.51 & 80.74 & \textbf{81.69} \\
\midrule
\multirow{3}{*}{JNLPBA} & Precision & 69.42 & - & 71.35 & 70.91 & \textbf{71.39} \\
& Recall & 75.99 & - & 75.74 & 76.34 & \textbf{79.06} \\
& F1 & 72.55 & 70.09 & 73.48 & 73.52 & \textbf{75.03} \\
\bottomrule
\end{tabular}
\captionof{table}{Precision, recall, and F1 scores of our proposed BiLM pretrained NER model (last column) and recent state-of-the-art models. We use the CNN-BiLSTM-CRF architecture for our NER model in all the experiments. Source of benchmark performance scores of datasets are: NCBI-disease: \citet{leaman2016taggerone}; BC5CDR: \citet{Li2015HITSZCDRSF}; JNLPBA: \citet{GuoDong:2004:EDK:1567594.1567616}; \textbf{MTM-CW} was proposed in \citet{Wang2018CrosstypeBN}; \textbf{FFN (Feed-forward network)} was proposed in \citet{crichton2017neural}; \textbf{BiLSTM} was proposed in \citet{doi:10.1093/bioinformatics/btx228}. The performance scores for these NER models are referred from \citet{Wang2018CrosstypeBN}. 
}
\label{table:results}
\end{table}

\subsection{Performance on Benchmark Datasets} We compare our proposed BiLM pretrained NER model with state-of-the-art NER systems such as the multi-task models of~\citet{crichton2017neural},~\citet{Wang2018CrosstypeBN}, and pretrained embedding based method of~\citet{doi:10.1093/bioinformatics/btx228}. We show the precision, recall, and F1 scores of the models for all the above datasets in Table~\ref{table:results}. From the results, we see that the approach of BiLM pretraining obtains the maximum F1 score for all the datasets. For NCBI-disease dataset, the F1 score of our model is $87.34\%$, which is an absolute improvement of $1.20\%$ over the multi-task learning method of~\cite{Wang2018CrosstypeBN}, in which they train the NER model jointly on all the datasets combined together. Similarly, for other datasets, we can see that our proposed approach outperforms other benchmark systems by a significant margin. We want to mention here that our model was trained only on the provided data for a particular dataset compared with the multi-task learning methods, which require a collection of labeled data to improve their performance. This also highlights the importance of doing pretraining of the model weights as this can improve their generalization ability on the test set.
\begin{table}[t]
\small
\centering
\begin{tabular}{l l r r r r}
\toprule
Dataset & Metric & No pretrain & 
$LM_f$ pretrain & $LM_b$ pretrain & BiLM pretrain \\
\midrule
\multirow{3}{*}{NCBI-disease} & Precision & 84.38 & 84.62 & 84.75 & \textbf{86.41} \\
& Recall & 87.37 & 87.89 & 88.00 & \textbf{88.31} \\
& F1 & 85.35 & 86.22 & 86.34 & \textbf{87.34} \\
\midrule
\multirow{3}{*}{BC5CDR} & Precision & \textbf{88.95} & 88.67 & 88.12 & 88.10 \\
& Recall & 88.64 & 89.28 & 89.41 & \textbf{90.60} \\
& F1 & 88.79 & 88.97 & 88.76 & \textbf{89.28} \\
\midrule
\multirow{3}{*}{BC2GM} & Precision & 81.40 & \textbf{82.00} & 81.04 & 81.81 \\
& Recall & 79.89 & 80.56 & 80.12 & \textbf{81.57} \\
& F1 & 80.62 & 81.27 & 80.58 & \textbf{81.69} \\
\midrule
\multirow{3}{*}{JNLPBA} & Precision & 71.23 & 70.51 & 71.00 & \textbf{71.39} \\
& Recall & 76.52 & 77.11 & 76.98 & \textbf{79.06} \\
& F1 & 73.78 & 73.66 & 73.87 & \textbf{75.03} \\
\bottomrule
\end{tabular}
\captionof{table}{Precision, recall, and F1 scores for different variations of our proposed model.}
\label{table:self-dataset}
\end{table}

\subsection{Model Variations Based on Weights Pretraining}
We also compare the performance of the following methods that are based on different parameter initialization strategies for the NER model.
\begin{itemize}
\item \textbf{No pretraining:} We randomly initialize the parameters of the NER model except word embeddings followed by supervised training.
\item $\boldsymbol{LM_f}$ \textbf{pretraining}: We initialize the parameters of the NER model using the forward language model weights. The parameters of backward LSTM, decoder, and CRF are randomly initialized.
\item $\boldsymbol{LM_b}$ \textbf{pretraining:} We initialize the parameters of the NER model using the backward language model weights. The parameters of forward LSTM, decoder, and CRF are randomly initialized.
\item $\boldsymbol{BiLM}$ \textbf{pretraining:} In this, the parameters of the NER model are initialized using the bidirectional language model weights. The parameters of the decoder and CRF are randomly initialized.
\end{itemize}

We show the results of the above variations in model pretraining in Table~\ref{table:self-dataset}. Our model gives an absolute improvement of around $2\%$ and $0.5\%$ in F1 score on NCBI-disease and BC5CDR dataset respectively over the model with no pretraining. We note that for all the datasets, $LM_f$ pretraining and $LM_b$ pretraining also gives an improvement over no pretraining. From the results, we also observe that the BiLM pretraining achieves better F1 score and precision in comparison to $LM_f$ pretraining and $LM_b$ pretraining, thus highlighting the importance of performing language modeling in both directions.

\subsection{Model Studies}
Next, we plot the precision-recall curve, convergence rate, and learning curve to gain additional insights about the NER model with BiLM pretraining and compare it with the randomly initialized model.

\begin{figure}[t]
\centering
\begin{minipage}{.5\textwidth}
\centering\includegraphics[scale=0.3, width=7.5cm, height=6cm]{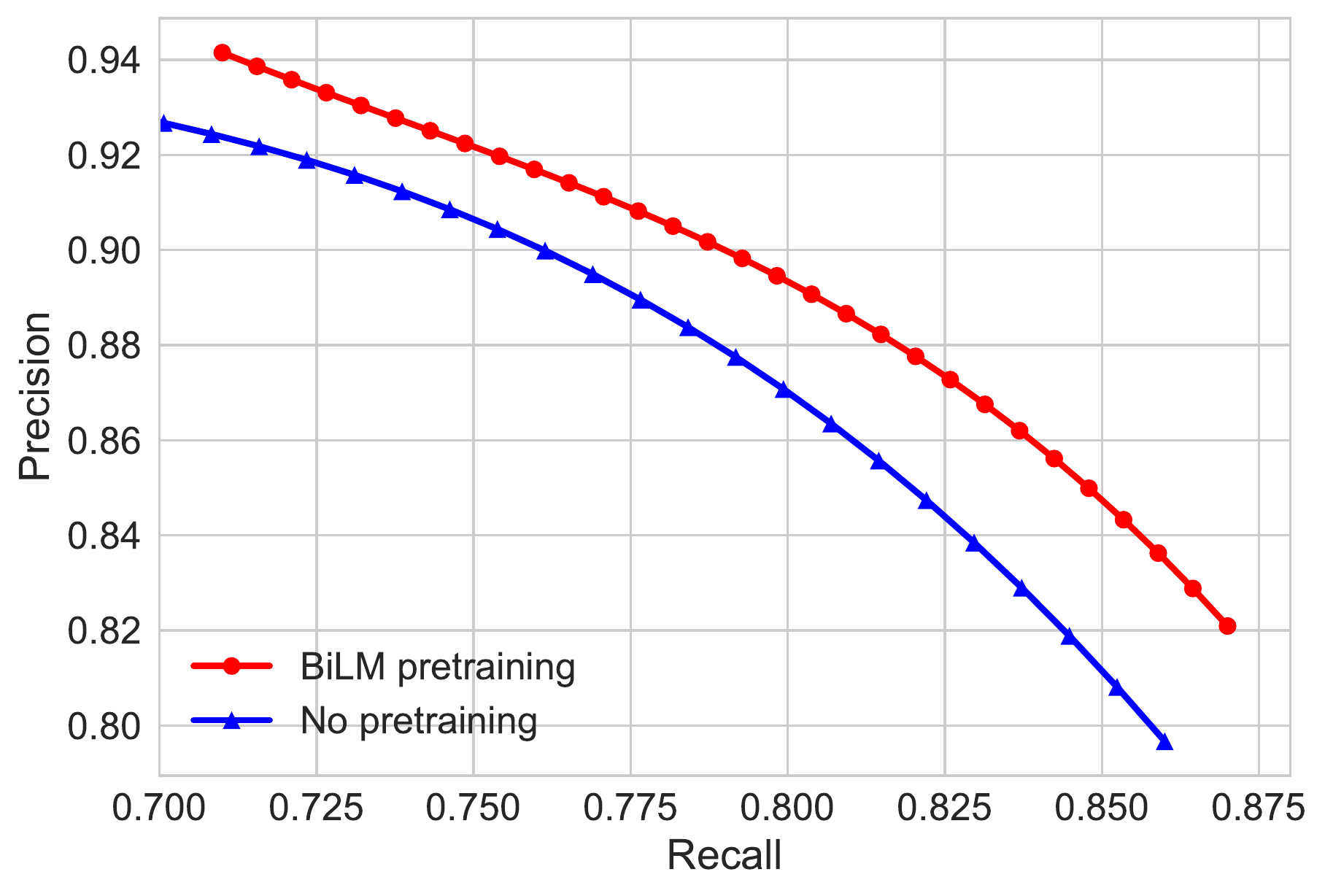}
  \subcaption{NCBI-disease dataset}\label{fig:ncbi_pr_curve}
\end{minipage}%
\begin{minipage}{.5\textwidth}
\centering\includegraphics[scale=0.3, width=7.5cm, height=6cm]{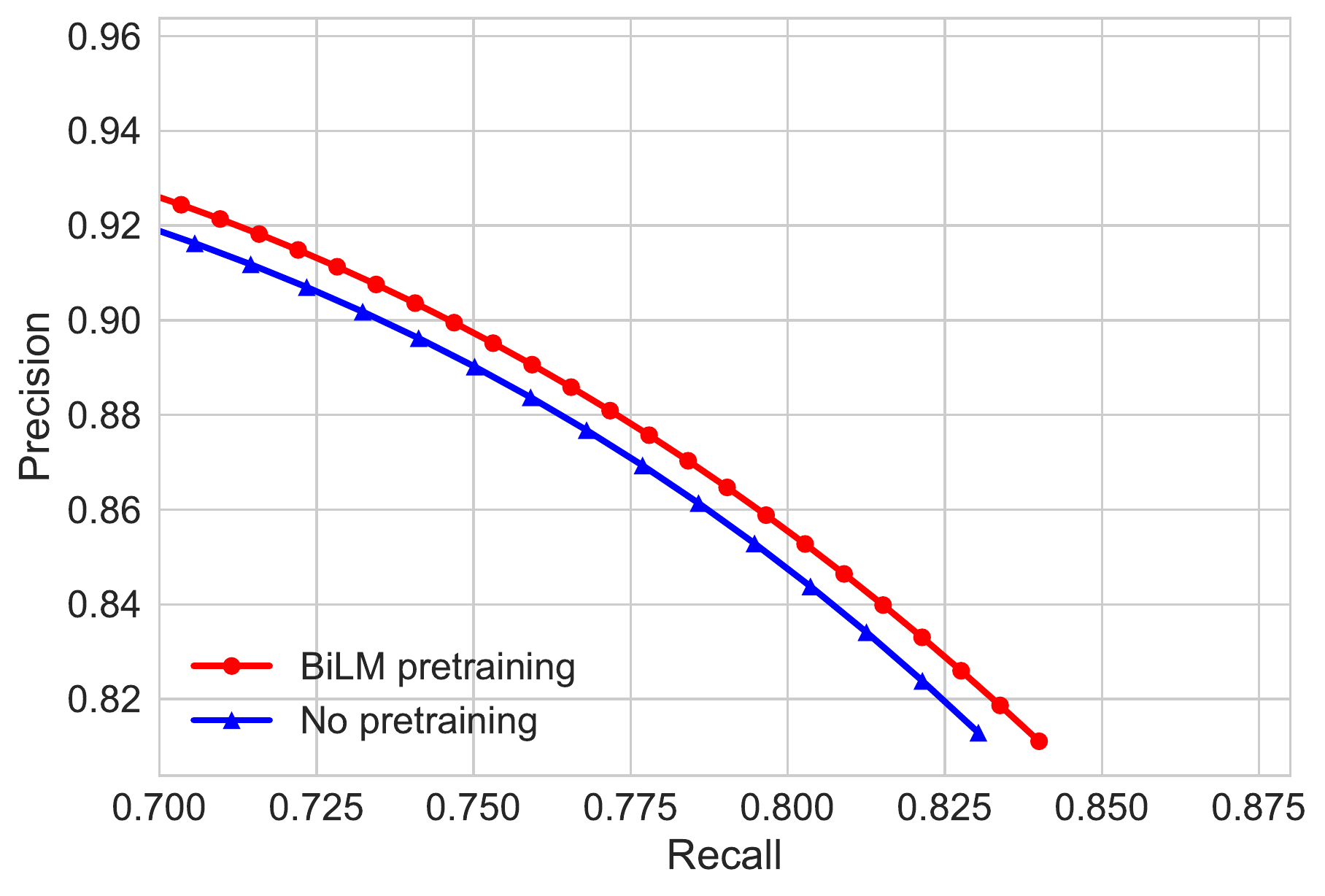}
  \subcaption{BC5CDR dataset}\label{fig:cdr_pr_curve}
\end{minipage}
\caption{Smoothed precision-recall curves for models with BiLM pretraining and no pretraining. Best viewed in color.}
\end{figure}

\subsubsection{Precision-Recall Curve} 
In Figure~\ref{fig:ncbi_pr_curve} and~\ref{fig:cdr_pr_curve}, we plot the smoothed precision-recall curve for NCBI-disease and BC5CDR datasets. From both the plots, we see that the BiLM pretrained NER model is always optimal as its area under the precision-recall curve is always more than that of a randomly initialized NER model.

\begin{figure}[t]
\centering
\begin{minipage}{.5\linewidth}\centering
  \includegraphics[scale=0.33, width=7.5cm, height=6cm]{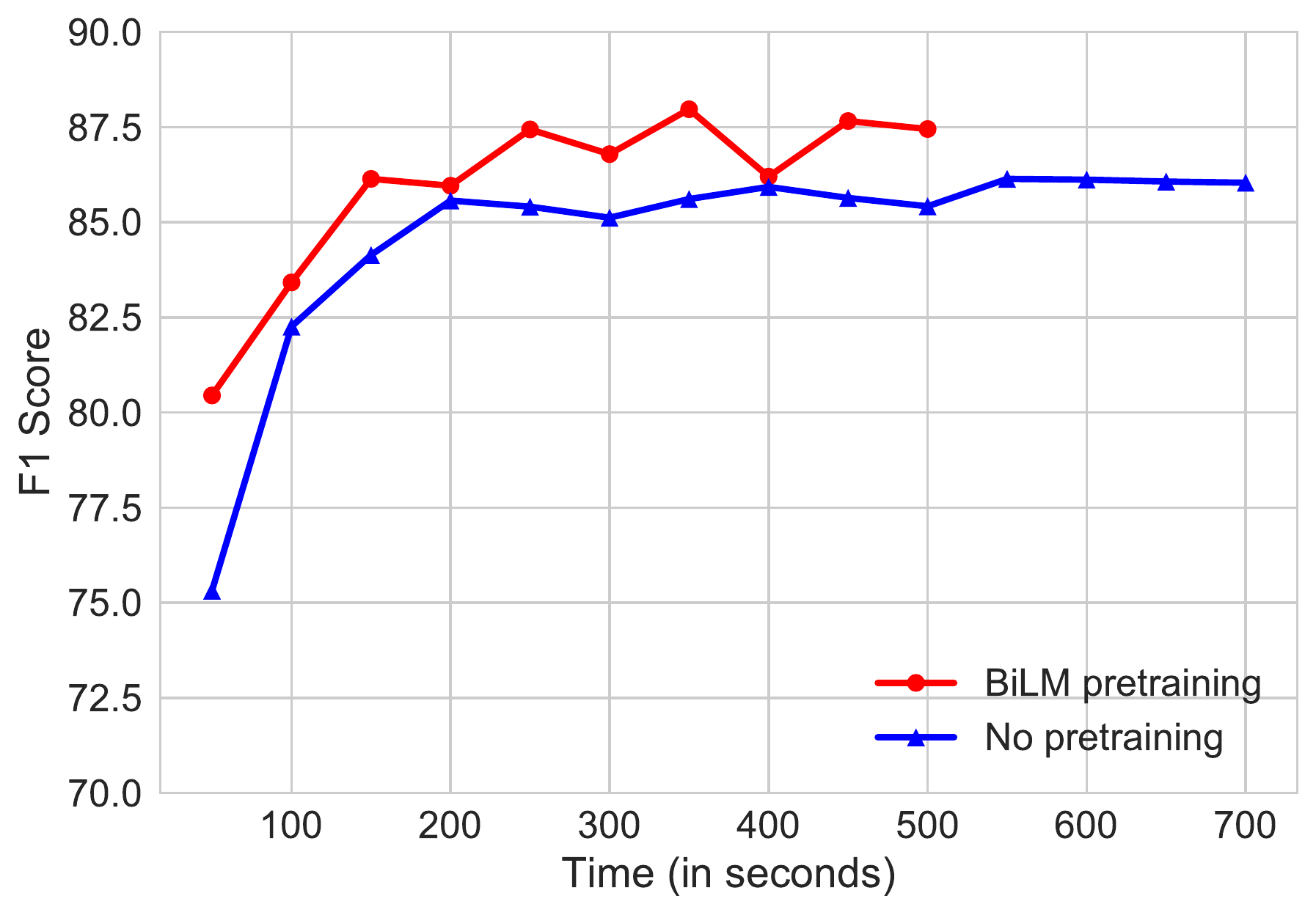}
  \subcaption{NCBI-disease dataset}
  \label{fig:ner_model_ncbi}
\end{minipage}%
\begin{minipage}{.5\linewidth}\centering
  \includegraphics[scale=0.33, width=7.5cm, height=6cm]{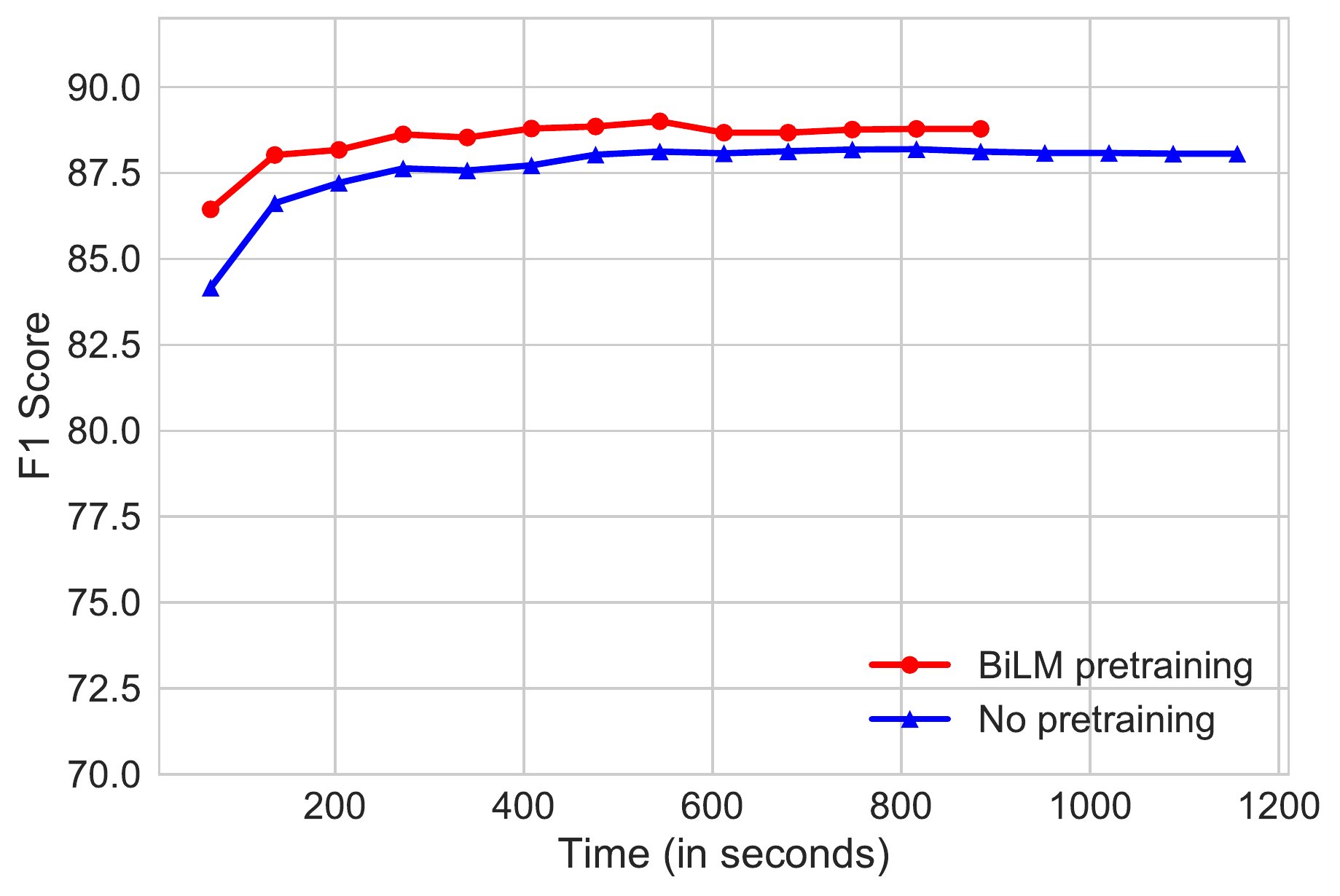}
  \subcaption{BC5CDR dataset}
  \label{fig:ner_model_cdr}
\end{minipage}
\caption{F1 score versus time taken for training to converge for models with BiLM pretraining and no pretraining. Best viewed in color.}
\end{figure}

\subsubsection{Rate of Convergence} 
We monitor the overall clock time and the time taken per epoch required for the two models to converge. We follow the same training process as outlined above. A typical run for both the models on NCBI-disease and BC5CDR dataset is shown in Figure~\ref{fig:ner_model_ncbi} and~\ref{fig:ner_model_cdr} respectively. For NCBI-disease dataset, the model with BiLM pretraining converges in 10 epochs ($\approx 500$s) compared with the model with no pretraining, which typically converges in 14 epochs ($\approx 700$s). We observe a similar trend in the BC5CDR dataset where BiLM pretraining results in convergence in 11 epochs ($\approx 900$s) whereas no pretraining takes around 17 epochs ($\approx 1150$s). Thus, in terms of total time taken, we observe that pretraining using BiLM weights results in faster convergence by about $28\mbox{-}35\%$ compared with random parameter initialization setting. We also see that BiLM pretraining results in a better F1 score from first epoch onwards for both the datasets.%

\subsubsection{Learning Curve} 
\begin{figure}[t]
\centering
\begin{minipage}{.5\textwidth}
  \centering
  \includegraphics[scale=0.3, width=7.5cm, height=6cm]{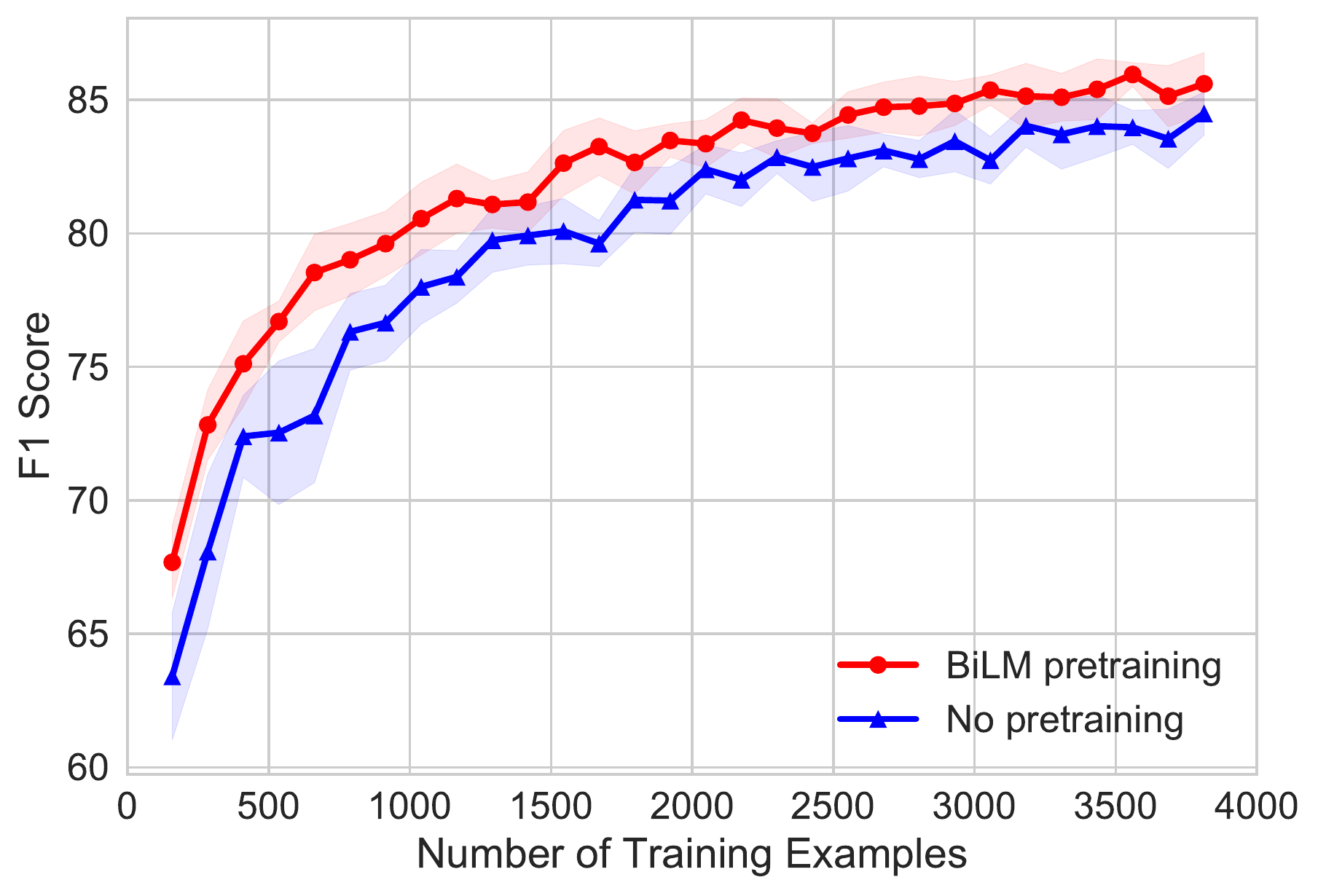}
  \subcaption{NCBI-disease dataset}
  \label{fig:ncbi_learning_curve}
\end{minipage}%
\begin{minipage}{.5\linewidth}
  \centering
  \includegraphics[scale=0.3, width=7.5cm, height=6cm]{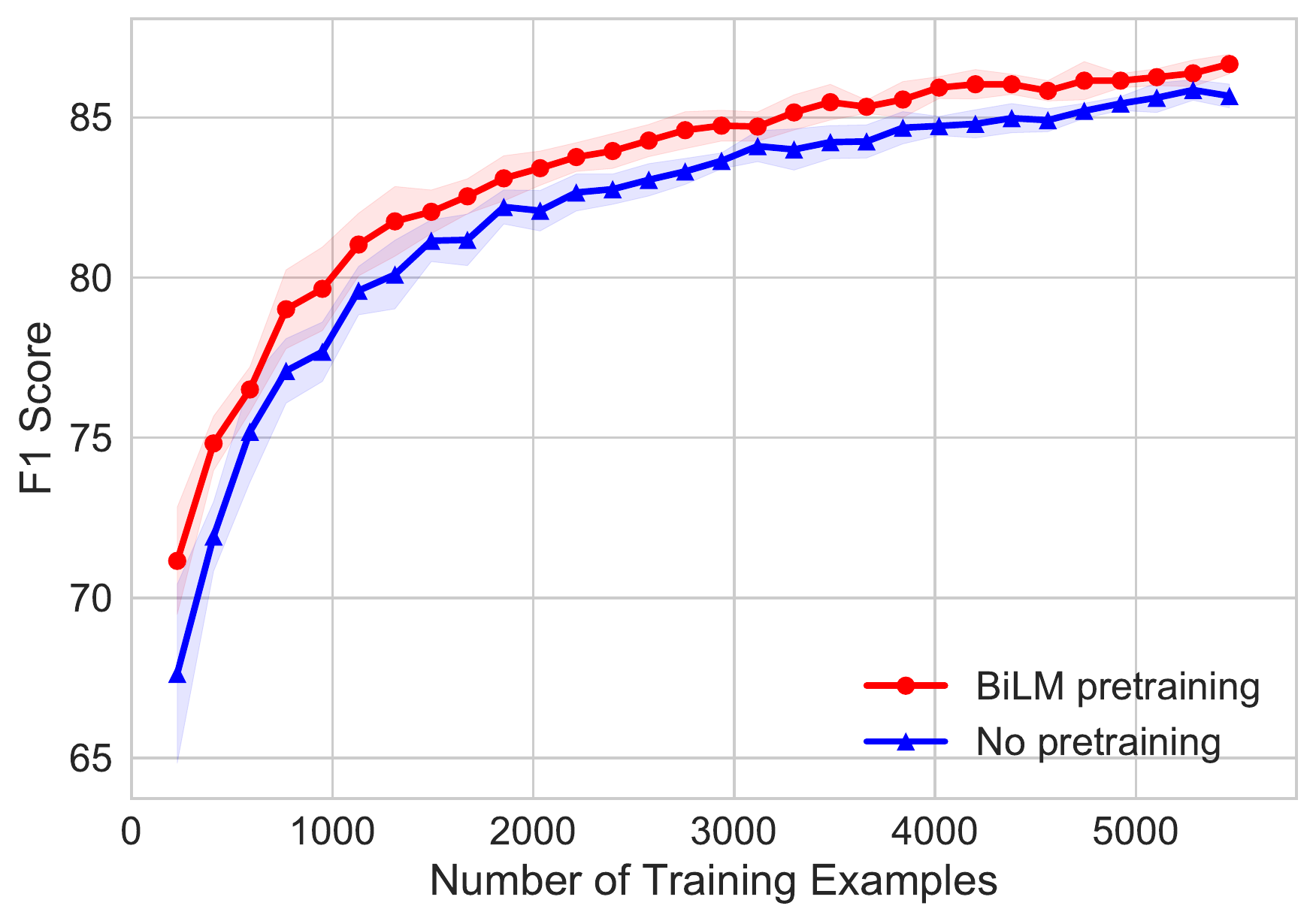}
  \subcaption{BC5CDR dataset}
  \label{fig:cdr_learning_curve}
\end{minipage}
\caption{F1 score versus increasing number of training examples for models with BiLM pretraining and no pretraining. Best viewed in color.}
\end{figure}

In this setup, we analyze the F1 score of both the models by feeding them with an increasing number of examples during the training process (learning curve). The learning curve for both the models on NCBI-disease and BC5CDR datasets is shown in Figure~\ref{fig:ncbi_learning_curve} and~\ref{fig:cdr_learning_curve} respectively. We can see that the BiLM pretrained model is always optimal (achieves higher F1 score) for any setting of the number of training examples.

\subsection{Case Study on NCBI-Disease Dataset} 
We will now discuss qualitative results of the BiLM pretrained NER model on NCBI-disease dataset. The NCBI-disease dataset consists of abstracts from medical research papers, which are written in a technical language and contains many complex entity names. 
In the NCBI-disease dataset, combined training and development set contains $1,902$ unique mentions of disease entities. In its test set, there are 423 unique occurrences of disease names and the BiLM pretrained NER model is able to correctly predict $365$ such diseases. Some examples of the longer disease names that are hard to recognize but our approach is able to correctly predict are ``\emph{sporadic breast , brain , prostate and kidney cancer}," ``\emph{deficiency of the ninth component of human complement}," ``\emph{von hippel - lindau ( vhl ) tumor}," and ``\emph{deficiency of the lysosomal enzyme aspartylglucosaminidase}." 

Among the $423$ unique mentions of diseases in the test set, $232$ of them are unseen in the combined training and development set. Our model was able to correctly predict around $120$ unseen disease entities in the test set. Some examples of unseen disease entities that are correctly predicted are ``\emph{deficiency of the lysosomal enzyme aspartylglucosaminidase}," ``\emph{campomelic - metaphyseal skeletal dysplasia}," ``\emph{atrophic benign epidermolysis bullosa}," and ``\emph{ectopic intracranial retinoblastoma}." This can be attributed to the improved modeling of the relationship among context words during bidirectional language modeling pretraining step. Some examples of the disease entities where our model fails are ``\emph{bannayan - zonana ( bzs ) or ruvalcaba - riley - smith syndrome}," ``\emph{very - long - chain acyl - coenzyme a dehydrogenase deficiency}," ``\emph{vwf - deficient}," and ``\emph{diffuse mesangial sclerosis}." From these examples, we see that the model makes an incorrect prediction when the disease entities have longer names, which may also contain abbreviations.

\section{Related Work} \label{sec:related_word}
Traditionally, researchers have worked on carefully designing hand-engineered features to represent a word such as the use of parts-of-speech (POS) tags, capitalization information, use of rules such as regular expressions to identify numbers, use of gazetteers, etc. A combination of supervised classifiers using such features was used to achieve the best performance on CoNLL-2003 benchmark NER dataset~\citep{florian2003named}.~\citet{Lafferty:2001:CRF:645530.655813} popularized the use of graphical models such as linear-chain conditional random fields (CRF) for NER tasks. Among the early approaches of NER systems in the biomedical domain include ABNER~\citep{settles2004biomedical}, BANNER~\citep{leaman2008banner}, and GIMLI~\citep{Campos2013}, which used a variety of lexical, contextual, and orthographic features as input to a linear-chain CRF.

The next generation of methods involves neural networks as they can be trained end-to-end using only the available labeled data without the need of manual task-specific feature engineering. In their seminal work,~\citet{collobert2011natural} trained window-based and sentence-based models for several NLP tasks and demonstrated competitive performance. For NER task on newswire texts,~\citet{huang2015bidirectional} uses word embeddings, spelling, and contextual features that are fed to a BiLSTM-CRF model. To incorporate character features,~\citet{lample2016neural} applies BiLSTM while ~\citet{chiu2016named,ma2016end} applies CNNs on character embeddings respectively. For biomedical NER task,~\citet{wei2016disease} combines the output of BiLSTM and traditional CRF-based model using an SVM classifier.~\cite{e19060283} experiments with character-level and word-level BiLSTM for the task of drug NER.~\citet{doi:10.1093/bioinformatics/btx228} investigates the effect of pretrained word embeddings on several biomedical NER datasets.

Pretraining the neural network model parameters using transfer learning has been widely studied and has shown to improve results in a variety of tasks such as deep autoencoders for dimensionality reduction~\citep{hinton2006reducing}, computer vision~\citep{erhan2010does}, text classification~\citep{dai2015semi,howard2018acl}, machine translation~\citep{ramachandran2017pretraining,ye2018pretrained}, and question answering~\citep{min2017acl}.

For sequence tagging tasks, supervised transfer learning to pretrain the model from the weights of another model that was trained on a different labeled dataset has been applied to domain adaptation tasks~\citep{qu2016named}, de-identification of patient notes~\citep{lee18transfer}, NER task in tweets~\citep{von2017transfer}, and biomedical NER~\citep{giorgi2018transfer,wang2018label}. In contrast, we pretrain the weights of the NER model from a language model that is trained on unlabeled data and thus removing the hard dependency on the availability of larger labeled datasets for pretraining.

\section{Conclusion}\label{sec:conclusion}
In this paper, we present a transfer learning approach for the task of biomedical NER. In our NER model, we use CNNs with different filters widths to extract character features and a word-level BiLSTM for sequence modeling which takes both word embeddings and character features as inputs. We pretrain the NER model weights using a BiLM such that the architectures of both the models are same except for the top decoder layer. The BiLM is trained in an unsupervised manner using only the unlabeled data.

We show that such pretraining of the NER model weights is a good initialization strategy for the optimizer as it leads to substantial improvements in the F1 scores for four benchmark datasets. Further, to achieve a particular F1 score, pretrained model requires less training data compared with a randomly initialized model. A pretrained model also converges faster during model fine-tuning. We also observe gains in the recall score for both seen and unseen disease entities.

For future work, we plan to train bigger sized language models on large collections of medical corpora and use it for providing additional features to the NER model so that it can incorporate wider context while training. We also plan to use external medical knowledge graphs to further improve the NER model's performance.

\acks{}
This work was supported by a generous research funding from CMU, MCDS students grant. We would also like to thank the anonymous reviewers for giving us their valuable feedback that helped to improve the paper.

\vskip 0.2in
\bibliography{main}

\end{document}